# Self-driving scale car trained by Deep reinforcement Learning

Qi Zhang
North China University of Technology

Tao Du
North China University of Technology

Changzheng Tian
North China University of Technology

## ABSTRACT
Self-driving scale car trained by Deep reinforcement Learning

This paper considers the problem of self-driving algorithm based on deep learning. This is a hot topic because self-driving is the most important application field of artificial intelligence. Existing work focused on deep learning which has the ability to learn "end-to-end" self-driving control directly from raw sensory data, but this method is just a mapping between images and driving. We prefer deep reinforcement learning to train a self-driving car in a virtual simulation environment created by Unity and then migrate to reality. Deep reinforcement learning makes the machine own the driving descision-making ability like human. The virtual to realistic training method can efficiently handle the problem that reinforcement learning requires reward from the environment which probably cause cars damge. We have derived a theoretical model and analysis on how to use Deep Q-learning to control a car to drive. We have carried out simulations in the Unity virtual environment for evaluating the performance. Finally, we successfully migrate te model to the real world and realize self-driving.

## 1. INTRODUCTION
The automotive industry is a special industry. In order to keep the passengers' safety, any accident is unacceptable. Therefore, the reliability and security must satisfy the stringent standard. The accuracy and rubustness of the sensors and algorithms are required extremely precision in the proccess of self-driving vehicles. On the other hand, self-driving cars are products for the average consumers, so the cost of the cars need be controled. High-precision sensors[1] can improve the accuracy of the algorithms but very expensive. This is a difficult contradiction need to solve.

Recently, the rapid development of artificial intelligence technology, especially the deep learning, has made a major breakthrough in the fields such like image recognition and intelligent control. Deep learning techniques, typically such as convolutional neural networks, are widely used in various types of image processing, which makes them suitable for self-driving applications. The researchers use deep learning to build end-to-end deep learning self-driving car whose core is learning through the neural network under supervised, then get the mapping relationship, finally achieve a pattern-replicating driving skills [2]. While end-to-end driving is easy to scale and adaptable, it has limited ability to handle long-term planning which involves the nature of imitation learning[3,4]. We perfer to let scale cares learn how to drive on their own than under human's supervision. Because there are many problems of this replication pattern, especially on the sensor. The traffic accidents of Tesla are caused by the failure of the perceived module in a bright light environment. Deep reinforcement learning can make appropriate decisions even some modules fail in working[5].

This paper focus on the issue of self-driving based on deep riforencement learning, we modify a 1:16 RC car and train it by double deep Q network. We use a virtual-to-reality process to achieve it, which means training the car in the virtual environment and testing in reality. In order to get a reliable simulation environment, we create a Unity simulation training environment based on OpenAI gym. We set a reasonable reward mechanism and modify the double deep Q-learning networks which makes the algrothm suitable for training a self-driving car. The car was trained in the Unity simulation environment for many episodes. At last, the scale car is able to learn a pretty good policy to drive itself and we successfully transfer the learned policy to the real world !

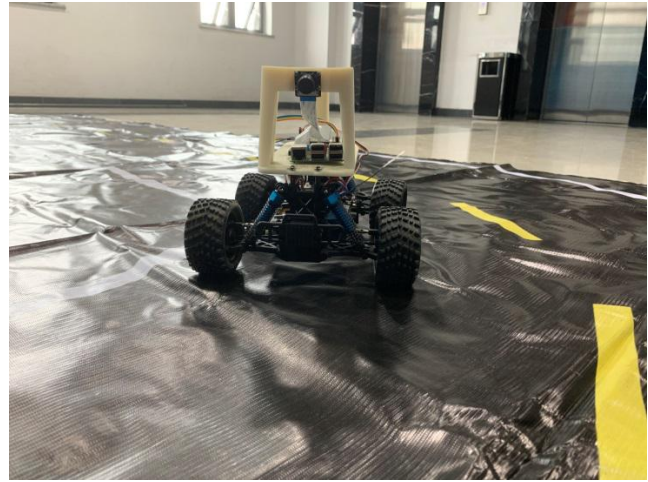

Figure 1: The reinforcement learning Donkey car based on DDQN.

## 2. RELATED WORK
Our aim is making a self-driving car trained by deep reinforcement learning. Right now, the most common methods to train the car to perform self driving are behavioral cloning and line following. On a high level, behavioral cloning works by using a convolutional neural network to learn a mapping between car images (taken by the front camera) and steering angle and throttle values through supervised learning. The other method, line following, works by using computer vision techniques to track the middle line and utilizes a PID controller to get the car to follow the line. Aditya Kumar Jain used CNN technology to complete the self-driving car with a camera[6]. Kaspar Sakmannti proposed a behavioral learning method [7], collecting human driving data through a camera, and then learning driving through CNN, which is a typical supervised learning.Kwabena Agyeman designed a car by linear regression versusblob tracking.However,these are the capabilities that under under manual intervention. We hope that cars can learn to drive by themselves,which is an intelligent way.

In 1989, Watkins proposed the noted Q-learning algorithm. The algorithm is mainly based on the Q table to record the state - the value of the action pair, each episode will update the state value. Mnih, Volodymyr, et al. In 2013, pioneered the concept of deep reinforcement learning [9], successfully applied in Atari games. In 2015 they also improved the model [10]. Two identically structured networks are used in DQN: Behavior Network and Target Network. Although this method improves the stability of the model, Q-Learning's problem of overestimating the value cannot be solved. To solve this problem, Hasselt proposed the Double Q Learning method, which is applied to DQN, which is Double DQN, DDQN [11]. The so-called Double Q Learning is to implement the selection of actions and the evaluation of actions with different value functions.

Recently, the use of virtual simulation techniques to train intensive learning models and then migrated to reality has been largely verified. OpenAI has developed a robotic arm called Dactyl [12] that trains AI robots in a virtual environment and finally applies them to physical robots. In the later research and exploration, the relevant personnel have been verified by the tasks of picking up and placing objects [13], visual servo [14], flexible movement [15], etc., all indicating their feasibility. In 2019, Luo, Wenhan, et al. proposed an end-to-end active target tracking method based on reinforcement learning, which trained a robust active tracker in a virtual environment through a custom reward function and environment enhancement technology.

From the above work, we can see that many of the visual autopilot algorithms learn through the neural network under the condition of supervised learning, get the mapping relationship, and then control. But this is not smart enough. Tesla's driverless accident is caused by perceived module failure in a bright light environment. Reinforced learning can do so, even in the event of failure of certain modules. Reinforcement learning makes it easier to learn a range of behaviors. Automated driving requires a series of correct actions to drive successfully. If only the data is annotated, the learned model is offset a bit at a time, and at the end it may be offset a lot, with devastating consequences. Reinforced learning can learn to automatically correct the offset. The key to a true autonomous vehicle is self-learning, and using more sensors does not solve the problem. It requires better coordination [16].

In this case, we use the algorithm of deep reinforcement learning to make our self-driving car.

## 3. Proposed method
### 3.1 Self-driving scale car

In autonomous vehicles, cars are often composed of traditional car-mounted sensor sensing systems, computer decision systems and driving control systems [17]. The function of the sensor sensing system is to capture surrounding environmental information and vehicle driving state, and provide information support for decision control. According to the scope of perception, it can be divided into environmental information perception and vehicle state perception. The environmental information includes roads, pedestrians, obstacles, traffic control signals and vehicle geographic location. Vehicle information includes driving speed, gear position, engine speed, wheel speed, and remaining. The amount of oil, etc. According to the implementation technology, it can be divided into ultrasonic radar, video acquisition sensor and positioning device [18].

In our desired experiment, we only need to use visual data as a sensing device. We use the RC car as a benchmark for retrofitting. The hardware used is:

Raspberry Pi (Raspberry Pi 3): This is a low-cost computer with a processing speed of 1.2 GHz and a memory of 1 GB. It is equipped with a customized version of the Linux system, supports Bluetooth, WIFI communication, and has rich support for i2c, etc. The agreement amount is GPIO port, which is the calculation brain for our auto-driving car.

PCA9685 (Servo Driver PCA 9685): Includes an i2 °C-controlled PWM driver with a built-in clock to drive the modified servo system.

Wide Angle Raspberry Pi Camera: The resolution is 2592 x 1944 and the viewing angle is 160 degrees. It is our only environmental sensing device, which is our eyes.

Other: For the sake of beauty, according to the design provided by the Donkey Car community, 3D printed a car bracket for carrying various hardware devices.

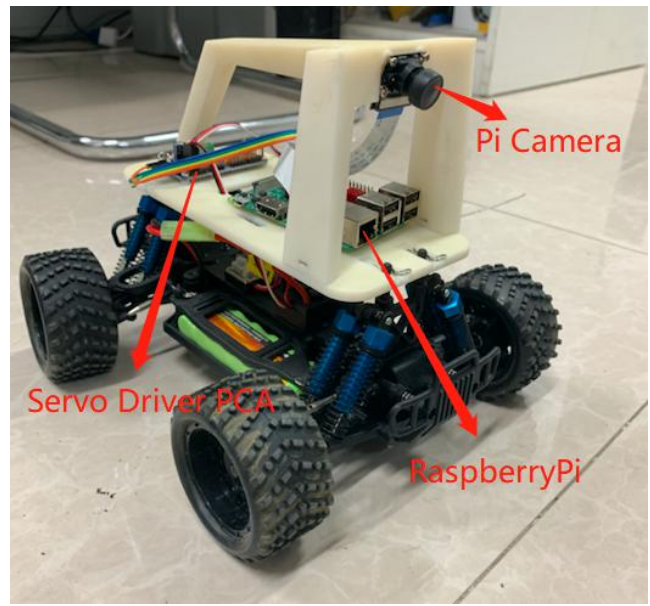

Figure 2:One 1:16 scale car.There is an opensource DIY self driving platform for small scale cars called donkeycar (visit donkeycar.com).

### 3.2 Environment require
#### 3.2.1 Donkey Car Simulator

The first step is to create a **high fidelity simulator** for Donkey Car. Fortunately, someone from the Donkey Car community has generously created a Donkey Car simulator in Unity. However, it is specifically designed to perform behavioral learning (i.e. save the camera images with the corresponding steering angles and throttle values in a file for supervised learning), but not cater for reinforcement learning at all. What we expected is an OpenAI gym like interface where we can manipulate the simulated environment through calling reset() to reset the environment and step(action) to step through the environment. We made some modifications to make it compatible with reinforcement learning.

Since we are going to write our reinforcement learning code in python, we have to first figure out a way to get python communicate with the Unity environment. It turns out that the Unity simulator created by Tawn Kramer also comes with python code for communicating with Unity. The communication is done through the Websocket protocol. Websocket protocol, unlike HTTP, allows two way **bidirectional communication between server and client**. In our case, our python "server" can push messages directly to Unity (e.g. steering and throttle actions), and our Unity "client" can also push information (e.g. states and rewards) back to the python server.

*3.2.2 Create a customized OpenAI gym environment for Donkey Car*

The next step is to create an OpenAI gym like interface for training reinforcement learning algorithms. For those of you who are have trained reinforcement learning algorithms before, you should be accustomed to the use of a set of API for the RL agent to interact with the environment. The common ones are reset(), step(), is_game_over(), etc. We can customize our own gym environment by extending the OpenAI gym class and implementing the methods above. The resulting environment is compatible with OpenAI gym. We can interact with the Donkey environment using the familiar gym like interface.

The environment also allows us to set frame_skipping and train the RL agent in headless mode(i.e. without Unity GUI).

Therefore,we have a virtual environment that we can use.

We take the pixel images taken by the front camera of the Donkey car, and perform the following transformations: 1.Resize it from (120,160) to (80,80) 2.Turn it into grayscale

3.Frame stacking: Stack 4 frames from previous time steps together 4.The final state is of dimension (1,80,80,4).

## 3.3 Algorithm

*3.3.1 The model of Reinforcement Learning*

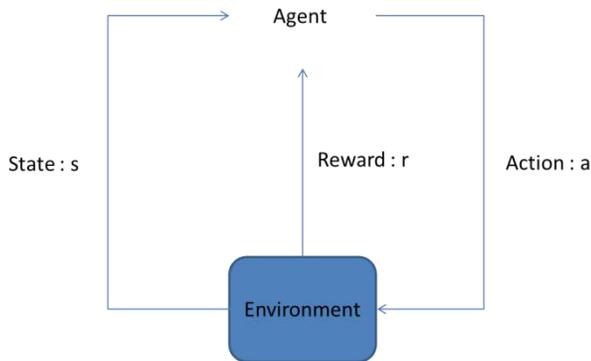

Figure 3:The process of reinforcement learning

Figure 3 shows the elements and processes of reinforcement learning. The agent takes action and interacts with the environment. The environment returns rewards and moves to the next state. Through multiple interactions, the agent gains experience and seeks the optimal strategy in experience. This interactive learning process is similar to the human learning style. Its main features are trial and error and delayed return. The learning process can be represented by the Markov decision process. The Markov decision process consists of triples <S, A, P, r>,where:

$$S = \{s_1, s_2, s_3 ...\}$$
$$A = \{a_1, a_2, a_3 ...\}$$
$$P^a_{ss'} = P[S_{t+1} = s' | S_t = s, A_t = a]$$
$$r = r(s,a)$$

S is a collection of all states; A is a collection of all actions; P is the state transition probability;$P^a_{ss'}$ means the transition probability when the agent takes action a and change state s to $s'$.The r is reward function,which means the reward of taking action a under state s.

The agent forms an interaction trajectory in each round of interaction with the environment($s_1,a_1,r_1,s_2,a_2,r_2...s_T,r_T,a_T$), The cumulative return at the state is

$$R_t = \gamma r_{t+1} + \gamma^2 r_{t+2} + \gamma^3 r_{t+3} + ... = \sum_{k=0}^{\infty} \gamma^k r_{t+k+1} \quad (1)$$

The $\gamma \in [0,1]$ is the discount coefficient of the return is used to weigh the relationship between current returns and long-term returns. The larger the value, the more attention is paid to long-term returns, and vice versa.

The goal of reinforcement learning is to learn strategies to maximize the expectations of cumulative returns:

$$\pi(a|s) = \underset{a}{\mathrm{argmax}}\, E[R] \quad (2)$$

In order to solve the optimal strategy, the value function and the action state value function are introduced to evaluate the advantages and disadvantages of a certain state and action. The value function is defined as follows:

$$V_\pi(s) = E_\pi\left[\sum_{k=0}^{\infty} \gamma^k r_{t+k+1} \middle| s_t = s\right]$$
$$= E_\pi[r_{t+1} + \gamma V(S_{t+1}) | s_t = s] \quad (3)$$

The action value function is defined as:

$$Q_\pi(s,a) = E_\pi\left[\sum_{k=0}^{\infty} \gamma^k r_{t+k+1} \middle| s_t = s, a_t = a\right] \quad (4)$$

Methods for solving value functions and action state value functions are based on table methods and approximation methods based on value functions [19]. Traditional dynamic programming, Monte Carlo and time difference (TD) algorithms are all table methods. The essence is to create a table of Q(s, a), behavioral state, and list as actions. The table is continuously updated by loop iteration calculation. value. When the state is relatively small, it is completely feasible, but when the state space is large, the traditional method is not feasible. Can you fit the state action value function with the approximating ability of the deep neural network to make Q(s, a) $\approx$ Q ( s, a, ø) has become the current research hot spot.

In 2013, deepmind highlighted the famous DQN algorithm [10], which opened a new era of deep reinforcement learning. The algorithm uses a convolutional neural network to approximate the state action value function, and uses the original pixels of the screen as input to directly learn the Atari game strategy. At the same time, using the experience replay mechanism [20], the training samples are stored in the memory pool, and each time a fixed amount of data is randomly sampled to train the neural network, the correlation between the training samples is eliminated, and the stability of the training is improved.

*3.3.2 Self-driving algorithm based on DDQN*

In the presence of a friendly reinforcement learning model training environment, we plan to use the strong learning algorithm as our control algorithm for automatic driving. For this we chose to use the DDQN algorithm because it has a relatively simple coding feature. Below we will introduce this method and how to apply it to the autopilot model.

In the DNQ algorithm, the author creatively proposed an approximate representation of the value function [9], successfully solving the problem that the problem is too large to be used. Among them, the state value function is introduced:

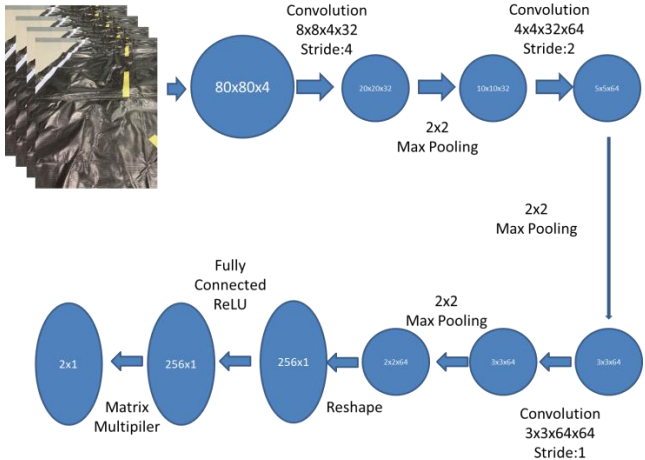

Figure 3:

And use neural networks to express state value functions. But it does not necessarily guarantee the convergence of the Q network, that is, we may not be able to get the Q network parameters after convergence. This will result in a poorly trained model. In order to solve this problem, DDQN [11] proposed by xx achieves the problem of eliminating overestimation by decoupling the selection of the target Q value action and the calculation of the target Q value.

DDQN (using the original name, abbreviated here) has two Q network structures like the DQN algorithm. In DDQN, it is no longer to find the maximum Q value in each action directly in the target Q network, but first find the action corresponding to the maximum Q value in the current Q network.

## 4. EXPERIMENT
### 4.1 Simulation

Essentially, we want our RL agent to base its output decision (i.e. steering) only on the **location and orientation of the lane lines** and **neglect everything else in the background**. However, since we give it the full pixel camera images as inputs, it might **overfit to the background patterns** instead of recognizing the lane lines. This is especially problematic in the real world settings where there might be undesirable objects lying next to the track (e.g. tables and chairs) and people walking around the track. If we ever want to transfer the learned policy from the simulation to the real world, we should get the agent to neglect the background noise and just focus on the track lines.

To address this problem, I've created a **pre-processing pipeline** to segment out the lane lines from the raw pixel images before feeding them into the CNN. The segmentation process is inspired by this excellent blog post. The procedure is described as follows:

1. Detect and extract all edges using **Canny Edge Detector**
2. Identify the straight lines through **Hough Line Transform**
3. Separate the straight lines into **positive sloped** and **negative sloped** (candidates for left and right lines of the track)
4. Reject all the straight lines that do not belong to the track utilizing slope information

The resulting transformed images consists of 0 to 2 straight lines representing the lane, illustrated as follows:

Figure 5:

We then took the segmented images, resize them to (80,80), stack 4 successive frames together and use it as the new input states. We trained DDQN again with the new states. The resulting RL agent was again able to learn a good policy to drive the car! With the setup above, I trained DDQN for around 100 episodes on a single CPU and a GTX 1080 GPU. The entire training took around 2 to 3 hours. As we can see from the video below, the car was able to learn a pretty good policy to drive itself!

Figure 6: The donkey car in the Unity Simulation.

Trained to get the model, the car learns to drive and stay in the center of the lane most of the time.

### 4.2 Simulation to Realty

We have customized a 3.5x4m simulation track from the merchants, and the track and Unity environment has a high degree of reduction, similar to the real life road (according to China's right-hand drive standard).

We modified the program to change the trained model input from Unity's output to the camera's real-time input. Then the program was configured in the Raspberry Pi and finally tested. The good news is that our car successfully followed the rules that he needed to follow, and operated on the right, and automatically turned.

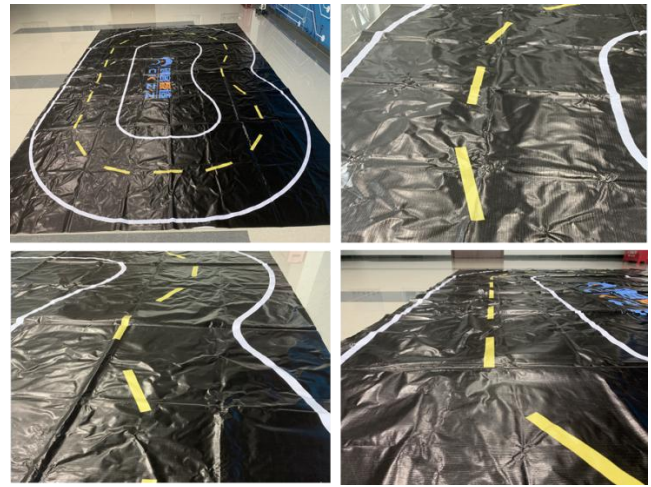

Figure 6: This is our road for Donkey car.

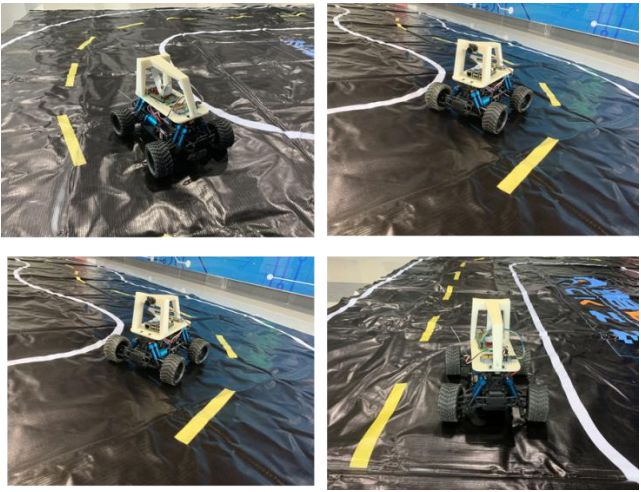

Figure 7: The trained car self-driving

We achieved the goal of autonomous driving, but the training time became very long after the image was processed, and the learned strategy would be accidentally unstable, especially in the case of a turn. The situation. After analysis, we found that this is to discard useful background information and line curvature information.

However, I've noticed that not only training took longer, but the learned policy was also less stable and the car wriggled frequently especially when making turns. I think this happened because we threw away useful background information and line curvature information.

In return, agents should not be too easy to over-fitting, or even summed up as invisible and real-world orbits.

Our papers demonstrate the use of deep reinforcement learning, coupled with training in the Unity simulator, and the resulting car is automatically driven within the tolerances.

## 5. CONCLUSION

In this article, we use a reinforcement learning algorithm to set up a model that can be automated with just one camera, plus virtual simulation training in Unity, and the resulting autonomous driving car completes the established driving goals. It is a very feasible way to conduct a strong learning and training through a virtual environment and then move to real life.